\documentclass{article}
\usepackage{spconf,amsmath,graphicx}
\usepackage{amsthm}
\usepackage{amsfonts}
\usepackage{bm}
\usepackage[ruled,vlined]{algorithm2e}
\usepackage{graphicx}

\def\R{{\mathbb R}}
\def\E{{\mathbb E}}
\usepackage{amssymb}%

\DeclareMathOperator*{\argmin}{arg\,min}

\usepackage{xcolor}

\usepackage{tikz}%
\usetikzlibrary{arrows}%
\tikzstyle{startstop} = [rectangle, rounded corners, minimum width=3em, minimum height=0.5em, text centered, draw=black, fill=red!30]%
\tikzstyle{process} = [rectangle, minimum width=5em, minimum height=1.5em, text centered, draw=black]%
\tikzstyle{arrow} = [ultra thick,->,>=stealth]%
\tikzstyle{int}=[draw, fill=blue!20, minimum size=2.5em]%
\tikzstyle{init} = [pin edge={to-,thick,black}]%

\title{ Compressive Learning for Semi-parametric Models }
\name{Michael P. Sheehan $^{\star}$  \qquad  Antoine Gonon $^{\dagger}$ \qquad Mike E. Davies $^{\star}$ \thanks{This work was supported by the ERC Advanced grant, project C-SENSE, (ERC-ADG-2015-694888). Mike. E. Davies is also supported by a Royal Society Wolfson Research Merit
Award.}}
\address{$^{\star}$ \texttt{\{firstname.lastname\}}@ed.ac.uk \quad IDCOM, University of Edinburgh, UK\\
$^{\dagger}$ antoine.gonon@ens-rennes.fr\quad Univ Rennes, F-35000 Rennes, France}
\begin{document}
\ninept
\maketitle
\begin{abstract}

In the compressive learning theory, instead of solving a statistical learning problem from the input data, a so-called sketch is computed from the data prior to learning. The sketch has to capture enough information to solve the problem directly from it, allowing to discard the dataset from the memory. This is useful when dealing with large datasets as the size of the sketch does not scale with the size of the database. In this paper, we reformulate the original compressive learning framework to explicitly cater for the class of semi-parametric models. The reformulation takes account of the inherent topology and structure of semi-parametric models, creating an intuitive pathway to the development of compressive learning algorithms. We apply our developed framework to both the semi-parametric models of independent component analysis and subspace clustering, demonstrating the robustness of the framework to explicitly show when a compression in complexity can be achieved.

\end{abstract}


\section{Introduction}
In the current era, it is common practice to have available very large datasets with millions of individual entries and hundreds of features. This poses a big challenge for large scale machine and statistical learning due to the fact computational and memory demands scale poorly with the dimensions of the dataset. The compressive learning framework \cite{gribonval2017compressive} was developed to tackle this issue and alleviate some of the complexity constraints. The principle of the framework is based on finding a compact representation, a so-called sketch, of the data prior to learning, such that enough information is preserved to minimise a form of risk associated to the learning problem. In general, the sketch does not scale with the size of the dataset but is driven by the complexity of the problem, making it amenable to large scale learning.  The framework has been successfully applied to various parametric models, including Gaussian mixture models  and $K$-means clustering \cite{keriven2017sketching}\cite{keriven2017compressive}, where the authors exploit explicit structural assumptions, residing in the probability space, to recover a risk from the sketch.

\par Semi-parametric models form an interesting class of models which are used extensively in the fields of machine learning, statistics and signal processing. One calls a statistical learning problem semi-parametric when the two following conditions are met: there are no parametric constraints on the data distribution and the learning problem can be entirely solved thanks to a statistic of the data distribution. For instance, one of the oldest semi-parametric models is principle component analysis (PCA), which can be solved by taking the eigenvalue decomposition of the covariance matrix of the data. The covariance matrix is an identifiable statistic sufficient to solve the PCA problem. The distinction between parametric and semi-parametric models is that we do not have access to a parametrized probability space for semi-parametric models, due to inherent topology and structure. Consequently, the original compressive learning framework does not naturally cater for semi-parametric models nor provide a pathway to design compressive learning algorithms.  

\par In this paper, we reformulate the original framework and apply it to semi-parametric models, leading to insights on creating compressive learning algorithms. Our main contribution is to recast the compressive learning framework to explicitly sketch the models identifiable statistics, exploiting structural assumptions in the statistic space, and show how this reformulation paves the way to creating compressive learning algorithms for both subspace clustering (SC) and independent component analysis (ICA).


\section{Background}
\vspace*{-1mm}
\subsection{Compressive Learning}
\vspace*{-1mm}
Let $x_1,x_2,\dots,x_N$ be independent and identically distributed samples from an unknown probability distribution $\pi$ on $(\mathcal{Z},\mathcal{B})$ where $\mathcal{Z} \subset\R^d$, some Euclidean space and $\mathcal{B}$ some Borel $\sigma$-field. Classically, $\pi$ is parametrized by some parameters denoted by $\theta\in\R^k$. A statistical learning problem can be formalised as follows: find a hypothesis $h^*$ from a hypothesis class $\mathcal{H}$ that best matches the probability distribution $\pi$ over the training collection $\{x_i\}_{i=1}^N$. Given a loss function $l:\mathcal{Z}\times\mathcal{H}\longmapsto\R$, this is equivalent to minimising the risk defined as
\noindent
\begin{equation}
    h^* = \argmin_{h\in\mathcal{H}}\mathcal{R}(\pi,h)=\argmin_{h\in\mathcal{H}}\E_{X\sim\pi}l(X,h).
\end{equation}
\noindent
Moreover, we define the model set associated to the hypothesis class as:
\noindent
\begin{equation}
    \mathfrak{S}_\mathcal{H}:=\{\pi\in\mathcal{P}(\mathcal{Z}):\exists h \in \mathcal{H}. \mathcal{R}(\pi,h)=0\}.
\end{equation}
\noindent
In other words, the set containing all distributions that are perfectly modeled by the hypothesis $h$. In practice, we generally do not have access to the true distribution $\pi$, so we instead minimise the empirical risk. As a consequence, this means we have to store all the data in memory. 

\par In compressive learning, we find a compact representation, or so-called sketch, that encodes some statistical properties of the data. Its size is ideally chosen relative to the intrinsic complexity of the problem, making it possible to work with arbitrarily large datasets while storing in memory an object of fixed size. Given a feature function $\Phi:\mathcal{Z}\longmapsto\R^m$, such that $\Phi$ is integrable with respect to any $\pi\in\mathcal{P}(\mathcal{Z})$, define a linear operator $\mathcal{A}:\mathcal{P}(\mathcal{Z})\longmapsto\R^m$ by
\noindent
\begin{equation}
\label{Eqn: sketch}
    \mathcal{A}(\pi):=\E_{X\sim\pi}\Phi(X).
\end{equation}
\noindent
We define our sketch in (\ref{Eqn: sketch}) to be the expectation of some features of the data distribution $\pi$. We want to choose $\mathcal{A}$ so that $\mathcal{A}(\pi)$ captures relevant statistical information of our data so that we'll be able to solve our learning problem from these observations directly. The goal of compressive learning is to therefore find a small $m\ll Nd$ that captures enough information to retrieve an estimated risk which is \textit{close} to the true risk with high probability \cite{gribonval2017compressive}. In practice, we use the empirical distribution and form an empirical sketch defined as 
\noindent
\begin{equation}
    \hat{y}=\mathcal{A}(\pi_N) \quad \text{where } \quad \pi_N:=\frac{1}{N}\sum_{i=1}^N\delta_{x_i}
\end{equation}
\noindent
denoting by $\delta_{x}$ the dirac distribution on $x$. Due to the law of large numbers,  $\lim_{N\rightarrow\infty}\mathcal{A}(\pi_N)=\mathcal{A}(\pi)$, the empirical sketch can be formed directly from our data.

\par
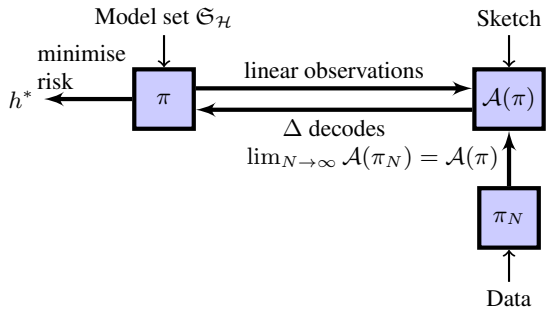
\begin{figure}[ht]
    \centering
    \begin{tikzpicture}[node distance=12.5em,auto,>=latex']
        \node [int,pin={[init]above:Sketch},ultra thick] (ske) {$\mathcal{A}(\pi)$};
        \node [int, pin={[init]above:Model set $\mathfrak{S}_\mathcal{H}$}, node distance = 14.5em,ultra thick] (mod) [left of= ske] {$\pi$};
        \node [node distance=6em,ultra thick] (solution) [left of= mod] {$h^*$};
        \node [int,pin={[init]below:Data}, node distance = 5em,ultra thick] (dat) [below of=ske] {$\pi_N$};
        \path[transform canvas={yshift=-1ex},->,ultra thick] (ske) edge node {$\Delta$ decodes} (mod);
        \draw[transform canvas={yshift=1ex},->,ultra thick] (mod) edge node {linear observations} (ske);
        \draw[->,ultra thick] (dat) edge[left] node {$\lim_{N\to\infty}\mathcal{A}(\pi_N)=\mathcal{A}(\pi)$} (ske);
        \draw[->,ultra thick] (mod) -- node [text width=4.25em,midway,above ] {\centering{minimise risk}} (solution);
    \end{tikzpicture}
    \vspace*{-3.5mm}
    \caption{A schematic diagram of the compressive learning framework}
    \label{fig: summary of compressive learning framework}
\end{figure}

 Once the sketch has been computed, one can discard the dataset $\{x_i\}_{i=1}^N$ from memory, reducing the memory complexity of the learning task. One can design a decoder $\Delta$ that exploits the structural assumptions of the model set $\mathfrak{S}_\mathcal{H}$ to recover a risk from the sketch. Consequently, we can find the best hypothesis $h^*$ by minimising the risk.  The sketching operator $\mathcal{A}$ and the decoder $\Delta$ form the pair $(\Delta,\mathcal{A})$ that define the compressive learning algorithm for a specific learning problem. A schematic diagram summarises the compressive learning framework in figure \ref{fig: summary of compressive learning framework}.
\noindent
\noindent
Gribonval and Keriven pioneered the method of compressive learning \cite{gribonval2017compressive} and successfully applied their framework to parametric models including Gaussian mixture models (GMM) and $K$-means clustering. In particular, they showed that the sketch algorithm reduces the memory complexity of GMM down to $m=\mathcal{O}(d\log_2(d))$ \cite{keriven2017sketching}, removing the dependency of the number of data points $N$. We say that $\mathcal{P}_\theta$ is a parametric model if it is a subset of the collection of all probability measures $\mathcal{M}$ on $(\mathcal{Z},\mathcal{B})$ which is fully described by a map $\theta\longmapsto \pi_\theta$ with $\theta$ ranging over $\Theta\subset\R^k$. In general, the parameter space is finite and the map $\theta\longmapsto \pi_\theta$ is smooth for parametric models. In fact, parametric models have the inherent property that a bijection exists between the map $\theta\longmapsto \pi_\theta$ which means that each $\theta$ corresponds to exactly one distribution. As we will see, this is not the case for semi-parametric models. \vspace*{-2.5mm}

\subsection{Semi-Parametric Models}
\vspace*{-1mm}
 Semi-parametric models contain sets that are a substantially large if not an infinite, subset of $\mathcal{M}$ on $(\mathcal{Z},\mathcal{B})$. These models are described by $\theta$, together with a function $g\in\mathcal{G}$, such that the model is specified by the set $\mathbb{P},\Theta,\mathcal{G}$ and the parametrization given by \cite{bickel1993efficient}:
 \noindent
\begin{equation}
    (\theta,g)\longmapsto \pi_{(\theta,g)} \quad\text{  for   } \,\,(\theta,g) \in \Theta \times \mathcal{G}.
\end{equation}
\noindent
We define a semi-parametric model by 
\noindent
\begin{equation}
\mathcal{P}_{(\theta,g)}:=\{\pi\in\mathcal{P}\mid \pi_{(\theta,g)}\text{, }\theta\in\R^k\text{, } g\in\mathcal{G}    \}.
\end{equation}
\noindent
In general, the map $(\theta,g)\longmapsto \pi_{(\theta,g)}$ is not bijective, and therefore one statistic corresponds to many distributions. This is the clear distinction between parametric and semi-parametric models. In most cases the function $g$ is not known, or is not sufficiently smooth to explicitly express in a concise parametrized way such that inference can be done. In numerous instances, we can use some statistics of the data which enable one to solve the semi-parametric task. As discussed, we can use the covariance matrix as a statistic to solve the PCA problem. Throughout this discussion we will term such statistics, which are used to solve the semi-parametric problem, as \textit{identifiable statistics.}
\vspace*{-1.5mm}
\section{Related Works}
\vspace*{-1mm}
 Recall that the covariance matrix acts as a identifiable statistic for the PCA problem i.e. the principal components can be found through the eigenspectrum of the covariance matrix. Given $\{x_i\}_{i=1}^N$, $x\in\R^d$, sampled from a probability distribution $\pi$ and covariance matrix $\Sigma_\pi\in \mathcal{S}\subset\R^{d\times d}$, we find the best $k$-dimensional subspace that best matches the data. This defines a hypothesis class for the PCA problem $\mathcal{H}=\{h\subset\mathcal{Z}\mid \text{ dim }h=k\}$ and a corresponding model set that is defined by the distributions that produce a covariance matrix of rank $k$:
 \noindent
\begin{equation*}
    \mathfrak{S}_\mathcal{H}=\{\pi\mid\text{ rank}(\Sigma_\pi)\leq k\}.
\end{equation*}
\noindent
Gribonval et. al. in \cite{gribonval2017compressive} show that one can take a sketch of the covariance matrix, $y=\mathcal{A}(\Sigma_\pi)$ , and decode the sketch to return an estimated risk by using a matrix completion algorithm that exploits the rankness of the covariance matrix. By doing so, one can reduce the complexity of the PCA task to $m=\mathcal{O}(kd)$. Sketched PCA does not succinctly fit into the compressive learning framework highlighted in figure \ref{fig: summary of compressive learning framework}. Firstly, notice that the sketched method, discussed above, encodes and decodes a statistic, which does not define a single probability distribution $\pi$ but infact an infinite number of distributions having the same covariance matrix. Secondly, we are not directly using structural assumptions on the model set to make the decoding step possible. Instead, we exploit structural assumptions from the intermediary set of identifiable statistics $\mathcal{S}$. In the next section we develop the framework to address these issues.
\vspace*{-1.5mm}
\section{Compressive Semi-Parametric Learning}
\begin{figure}[ht]
    \centering
     \includegraphics[width=0.95\linewidth]{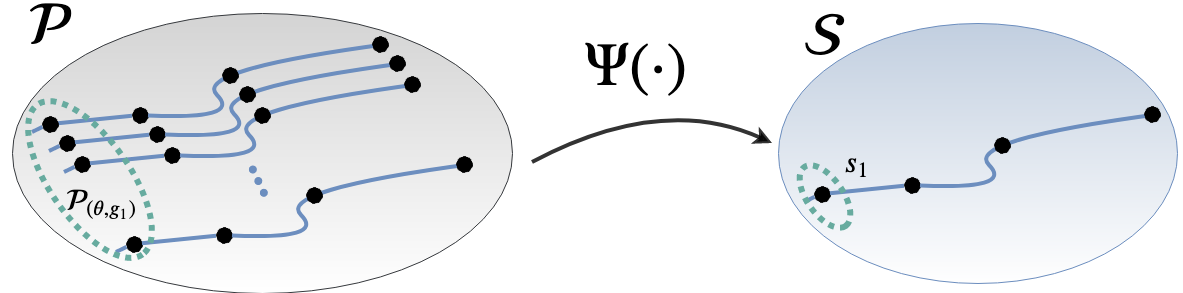}
     \vspace*{-3.mm}
    \caption{A schematic diagram of the probability equivalence class where many distributions collapse down to one point in the statistic set.}
    \label{fig:equivalence class}
\end{figure}

\vspace*{-1mm}
In section 3, we showed that compressive PCA does not succinctly fit the compressive learning framework nor provide any intuition how to create a compressive PCA algorithm. The covariance matrix corresponds to infinitely many data distributions and therefore it is almost impossible to decode a single distribution $\pi$.  Indeed, this is the case for all semi-parametric models. With an abuse of notation, let $\Sigma_\pi$ denote the identifiable statistic associated to an arbitrary semi-parametric model. An equivalence exists between distributions in the model set $\mathfrak{S}_\mathcal{H}$ and the set of identifiable statistics $\mathcal{S}$. Formally, let $\sim$ be the equivalence relation defined by
\noindent
\begin{equation}
    \{ \pi_a\sim\pi_b\mid \Sigma_{\pi_a}=\Sigma_{\pi_b}\}.
\end{equation}
\noindent As a result, there exists a many-to-one mapping, $\Psi:\mathcal{P}_{(\theta,g)}\mapsto \mathcal{S},$ that maps equivalence classes in $\mathcal{P}_{(\theta,g)}$ to the same point in $\mathcal{S}$. Both the equivalence class structure and the mapping are illustrated in Figure \ref{fig:equivalence class}.
\noindent
\noindent
Due to the equivalence class structure inherent in semi-parametric models, we lose the luxury of injectivity found in parametric models for the mapping $\Psi$. The consequence is that a single distribution cannot be decoded, and therefore the original framework does not cater explicitly for semi-parametric models. Below, we define a reformulation of the compressive learning framework to tackle such models and provide a pathway to develop compressive learning algorithms amenable to semi-parametric models. 
\vspace*{-2.5mm}
\subsection{Reformulated Framework}\label{subsection: reformulated framework}
\vspace*{-1mm}
We reformulate the framework by assuming that we know a statistic set $\mathcal{S}$ that can be used to define the risk function. That means that instead of having one risk function per distribution as before, here we have one risk function per equivalence class. This is possible when there exists a map $\Psi:\mathcal{P}_{(\theta,g)}\longmapsto \mathcal{S}$ satisfying 
\noindent
\begin{equation}
    \mathcal{R}(\pi,h)=\mathcal{R}(\Psi(\pi),h).
\end{equation}
\noindent
It turns out that the parameterization of the probability distributions is not needed anymore. Indeed, it suffices to have a parameterization of the statistic set to search for a sketch. Note that the size of the set $\mathcal{S}$ containing the statistics may be smaller than the size of the model set, as many probability distributions have the same statistic. In accordance, we define the new sketch as
\noindent
\begin{equation}
    \mathcal{A}(\Psi(\pi))=\E_{x\sim\pi}\Phi(X)
\end{equation}
\noindent
where $\mathcal{A}:\mathcal{S}\longmapsto\R^m$ is a linear operator on $s\in\mathcal{S}$ and $\Phi:\mathcal{Z}\longmapsto\R^m$ is a given feature function. As we are encoding a statistic from finite samples, the empirical sketch is defined as $\hat{y}:=\mathcal{A}(s_N)$, where $s_N$ is the empirical statistic computed through the samples. As ever, the law of large numbers apply, such that $\lim_{N\to\infty}\mathcal{A}(s_N)=\mathcal{A}(s)$. Once the sketch is formed, we use a decoder $\Delta_s$ that recovers a statistic $\hat{s}\in\mathcal{S}$ such that $\mathcal{R}(\hat{s},\cdot)$ and $\mathcal{R}(s,\cdot)$ are uniformly close. The decoder $\Delta_s$ is designed specifically to exploit the structural assumptions of the set $\mathcal{S}$. Consequently, we can find the best hypothesis $h^*$ by minimising the risk. Assuming that $\Sigma_\pi\in\mathcal{S}$ is our identifiable statistic associated with a semi-parametric model, a schematic diagram of the reformulated framework is highlighted in figure 3. 

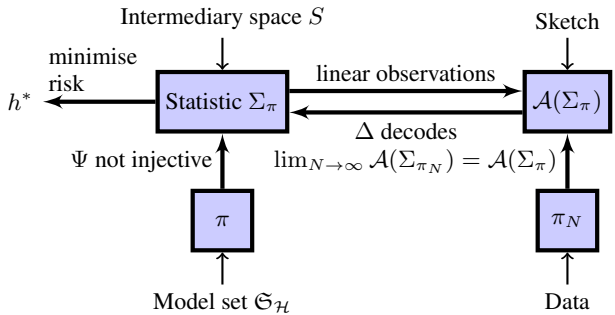
\begin{figure}[h!]
    \centering
    \begin{tikzpicture}[node distance=12.5em,auto,>=latex']
        \node [int, pin={[init]above:Intermediary space $S$},ultra thick] (sta) {Statistic $\Sigma_\pi$};
        \node [node distance=8.5em,ultra thick] (solution) [left of= sta] {$h^*$};
        \node [int, pin={[init]below:Model set $\mathfrak{S}_\mathcal{H}$}, node distance = 5em,ultra thick] (mod) [below of=sta] {$\pi$};
        \node [int,pin={[init]above:Sketch}, node distance = 14.5em,ultra thick] (ske) [right of=sta] {$\mathcal{A}(\Sigma_\pi)$};
        \node [int,pin={[init]below:Data}, node distance = 5em,ultra thick] (dat) [below of=ske] {$\pi_N$};
        \path[->,ultra thick] (mod) edge node {$\Psi$ not injective} (sta);
        \path[transform canvas={yshift=-1ex},->,ultra thick] (ske) edge node {$\Delta$ decodes} (sta);
        \draw[transform canvas={yshift=1ex},->,ultra thick] (sta) edge node {linear observations} (ske);
        \draw[->,ultra thick] (dat) edge[left] node {$\lim_{N\to\infty}\mathcal{A}(\Sigma_{\pi_N})=\mathcal{A}(\Sigma_\pi)$} (ske);
        \draw[->,ultra thick] (sta) -- node [text width=4.25em,midway,above ] {minimise risk} (solution);
    \end{tikzpicture}
    \label{fig: compressive SP framework}
    \vspace*{-3.5mm}
    \caption{A schematic diagram of the new compressive semi-parametric learning framework.}
\end{figure}
Our new formulation of the compressive learning framework provides a far more intuitive and explicit pathway enabling one to  identify statistics associated with a given semi-parametric model to create compressive learning algorithms. Furthermore, the framework allows one to explicitly design a decoder $\Delta_s$ by demonstrating the structural assumptions of semi-parametric models, where the original framework severely lacked. In the next section, we shall signify the importance of our reformulation by applying the  framework to two well known semi-parametric models. 
\vspace*{-1.5mm}
\section{Case Studies}
\vspace*{-1.5mm}
In this section we apply our compressive semi-parametric framework to two well known, yet complex, semi-parametric models of independent component analysis and subspace clustering. To be consistent and for comparison, we will use the notation $\Sigma_\pi$ to denote the identifiable statistic for each model.
\vspace*{-2mm}
\subsection{Compressive Independent Component Analysis}
\vspace*{-1mm}
We start the discussion with ICA, a semi-parametric model that decomposes data into hyperplanes of maximum independence via a linear transformation. For the sake of simplicity and brevity, we assume the data has identity covariance and zero mean, and therefore the task of ICA is to find an orthogonal matrix $Q$ such that $x=Qs$, where $s$ has statistically independent entries : $p(s)=\prod_{i=1}^d p_i(s_i)$. Each $p_i$ denotes the distribution of an independent component, and as we do not know the nature of the densities $p_i$ in advance, we cannot reduce the estimate done to a finite parameter set. Consequently, the estimation of $p_i$ is  non-parametric, and coupled with the parametric part of estimating the orthogonal matrix $Q$, results in ICA belonging to the class of semi-parametric models.
\par We resort to higher order statistics, specifically kurtosis, to solve the problem \cite{hyvarinen2000independent}. In general, kurtosis is a measure of independence for sources of different densities. Minimising the kurtosis of entry wise sources, maximises the independence of the system \cite{comon1994independent}. In our setting described, each point-wise kurtosis, defined by:
\noindent
\begin{equation}
   \Sigma_{ijkl}=\E[x_ix_jx_kx_l]-3,
\end{equation}
\noindent
forms a $4^{th}$ order kurtosis cumulant tensor $\Sigma_\pi\in\R^{d\times d\times d\times d}$. The goal of cumulant based ICA, is therefore to find an orthogonal transformation $Q$:
\noindent
\begin{equation}
    \Sigma_\pi \times_1 Q^T \times_2 Q^T \times_3 Q^T \times_4 Q^T
\end{equation}
\noindent
resulting in zero cross cumulants $\Sigma_{ijkl}=0$  $\forall ijkl\neq iiii $. Consequently, the sources will be independent and the cumulant tensor will be diagonal. The set of diagonal cumulant tensors can be defined as
\noindent
\begin{equation}
    \mathcal{D}:=\{\Sigma_\pi\mid {\Sigma_\pi}_{ijkl}=0 \quad \forall\, ijkl\neq iiii\}. 
\end{equation}
\noindent
By doing so, we can define the model set $\mathfrak{S}_\mathcal{H}$ of the ICA model:
\noindent
\begin{equation}
 \mathfrak{S}_\mathcal{H}:=\{\pi\mid \Sigma_\pi\times_1Q^T\times_2Q^T\times_3Q^T\times_4Q^T\in\mathcal{D}\},   
\end{equation}
\noindent
where $Q$ is the parameter of interest and $\times_j$ denotes the $j^{th}$ matrix-tensor product. 
\par The new formulation of compressive learning for semi-parametric models described in section \ref{subsection: reformulated framework} shows we must look for structural assumptions on the statistic set $\mathcal{S}\subset\R^{d\times d\times d \times d}$ to sketch $\Sigma_\pi$. In the case of ICA, the assumption that the cumulant tensor (formed from data $\{x_i\}_{i=1}^{N}$) can be \textit{diagonalised} by an orthogonal transformation, results in the solution living on a manifold, denoted $\mathcal{Q}$, of dimension $d^2$ compared to that of $d^4$ of the statistical set \cite{Sheehan2019CICA}. More precisely, it is sufficient to take $m=\mathcal{O}(d^2)$ random linear projections of $\Sigma_\pi$. A compressive ICA algorithm can be defined by the encoding-decoding pair $(\Delta_s,\mathcal{A})$:
\noindent
\begin{equation} \label{eq: linear operator and decoder for sketched ICA}
\begin{cases}
            \mathcal{A}:\Sigma\in\mathbb{R}^{d\times d\times d\times d} \mapsto \big(a_i^T\text{vec}(\Sigma)\big)_{i=1:m} \in \R^m \\
            
            \Delta_s(\hat{y}) = \argmin_{\substack{\hat{y}=\mathcal{A}(\Sigma )\\\Sigma\in\mathcal{Q}}}\rho_Q(\Sigma)
\end{cases}
\end{equation} 
\noindent
where $a_i\sim\mathcal{N}\big(0,\frac{1}{\sqrt{m}}\mathbf{I}_d\big)$ and $\rho_Q$ defines any independence contrast function defined over cumulant tensors \cite{comon1994independent}. 
\par
\begin{figure}[ht]
    \centering
    \includegraphics[width=0.95\linewidth]{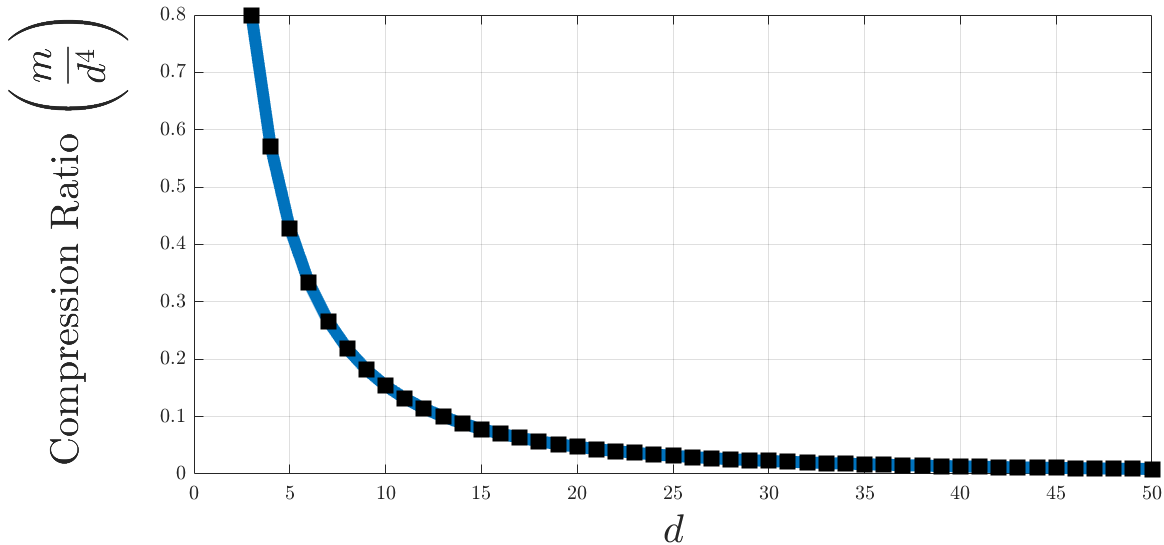}
    \vspace*{-3.5mm}
    \caption{Compressive ICA learning: A graph showing the compression ratio $\frac{m}{d^4}$ as $d$ grows.}
    \label{fig: ICA Compression Ratio}
\end{figure}

 Figure \ref{fig: ICA Compression Ratio} shows the ratio of compression $\frac{m}{d^4}$ as $d$ grows. The figure illustrates clearly that the framework has enabled us to identify a statistic that lives in a set with strong structural assumptions, that can be sketched to vastly reduce the order of memory complexity.
\noindent
\noindent

\vspace*{-3mm}
\subsection{Compressive Subspace Clustering}
The subspace clustering problem consists of finding the best union of subspaces that matches the given data $\{x_i\}_{i=1}^N\in\R^d$ \cite{vidal2011subspace}. It can be formalised by the hypothesis class
\noindent
\begin{equation}
    \mathcal{H}:=\{(h_1,h_2,\dots,h_n) \mid h_i\subset\R^d, \text{dim }h_i=d_i\},
\end{equation}
\noindent
which forms the corresponding model set
\noindent
\begin{equation}
    \mathfrak{S}_\mathcal{H}:=\{\pi=\sum_i\alpha_i\pi_{h_i}\mid\text{ rank}(\Sigma_{\pi_{h_i}})\leq d_i, \sum_i\alpha_i=1\}.
\end{equation}
\noindent
In the literature, we assume that the number of subspaces $n$ and the dimension $d_i$ of each subspace are known in advance, to make the problem well-posed. Subspace clustering can be thought of as a $n$-mixture model with data sampled from unknown  probability distributions $\pi_{h_i}$. As we do not know the form of these distributions, we can not reduce the estimate down to a finite parameter set. Similar to the ICA case, estimating $\pi_{h_i}$ is non-parametric, and coupled with the parametric form of the mixture coefficients $\alpha$, results in subspace clustering fitting into the semi-parametric class of models.
\par
\begin{figure}[ht]
    \centering
    \includegraphics[width=0.95\linewidth]{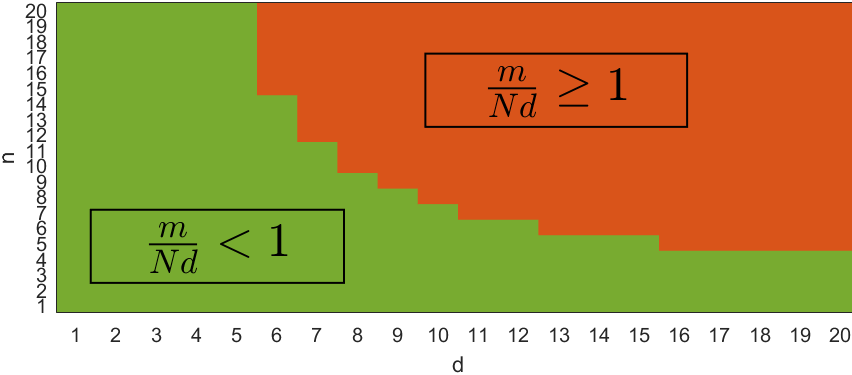}
    \vspace*{-3.5mm}
    \caption{Compressive GPCA learning: The compression ratio of the sketch size and the data length, $\frac{m}{Nd}$, as the model dimensions $n$ and $d$ grow. The data length is fixed at $N=10000000$. }
    \label{fig: GPCA Compression Ratio}
\end{figure}

 The subspace clustering problem can be solved through a generalised principle component analysis (GPCA) approach \cite{vidal2005generalized}. By assuming the data $\{x_i\}^N_{i=1}$ lies within a union of $n$ subspaces $S$, we denote by $\nu_{n,d}(x)$, or $\nu(x)$ for simplicity, the vectors having components equal to all the monomials of degree $n$ in the $d$ components of the data point $x$. For instance when $n=2$ and $d=3$, $\nu_{n,d}(x)=(x_1^2 \,x_1x_2\, x_1x_3\, x_2^2\, x_2x_3\, x^2_3)^T$. The embedded point $v_{n,d}(x)$  belongs to $\R^D$ with
\begin{equation}
    D:=\binom{n+d-1}{d-1}\leq d^n.
\end{equation}
For any union $S$ of $n$ subspaces $S_i$, we can find polynomials $(p_j)_{j=1...\bar{R}}$ that define the union of subspaces:
\begin{equation}\label{eq: characterization of a UoS by an algebraic variety}
    x\in \bigcup\limits_{i=1}^{n}S_i \Leftrightarrow  \bigwedge\limits_{j=1}^{\bar{R}}p_j(x)=0 \text{,}
\end{equation}
by computing the null space of the matrix $V^T=[\nu(x_1),\dots,\nu(x_N)]^T.$ Indeed, computing the null space of $V$ can be easily deduced by finding the eigendecomposition of the correlation matrix $\Sigma_\pi$ of the embedded data:
\begin{equation}
    \Sigma_\pi := \frac{1}{N}VV^T \in\R^{D\times D}.
\end{equation}
The correlation matrix of the Veronese embeddings is therefore the identifiable statistic associated to subspace clustering and we can therefore apply the compressive semi-parametric framework to it. As expected, the framework motivates us to seek structural assumptions of the statistic set $\mathcal{S}\subset\R^{D\times D}$. In the situation of GPCA, the correlation has rank $R$ between 1 and $D$ depending on the geometric makeup of the subspaces. In certain cases, the rank of correlation is in fact very small and therefore the degrees of freedom are far less than the dimensions of the statistic set $S$. In such situations, we know that only $\mathcal{O}(DR)$ measurements are needed to recover $\Sigma_\pi$ and therefore it is sufficient to take $m=\mathcal{O}(DR)$ rank-one projections of $\Sigma_\pi$ to enable stable recovery. A compressive GPCA algorithm can be defined by the encoding-decoding pair $(\Delta_s,\mathcal{A})$:

\begin{equation} \label{eq: linear operator and decoder for sketched gpca}
\begin{cases}
            \mathcal{A}:\Sigma\in\mathbb{R}^{D\times D} \mapsto (\text{trace}(a_ia_i^T\Sigma))_{1\leq i \leq m}\in\mathbb{R}^m \\
            
            \Delta_s(\hat{y}) = \argmin_{\substack{\Sigma \in\mathbb{R}^{D\times D} \\ \Sigma\succcurlyeq 0 \\  \hat{y}=\mathcal{A}(\Sigma)}} \|\Sigma\|_* 
\end{cases}
\end{equation}
Figure \ref{fig: GPCA Compression Ratio} shows a phase transition for the ratio of memory compression $\frac{m}{Nd}$ with respect to the dataset size $Nd$ as the dimension $d$ and number of subspaces $n$ grows, when the correlation matrix $\Sigma_\pi$ is of low rank $(R=0.05D)$. The green region shows when compression occurs, while the red region shows when compression is not possible in comparison to storing the whole data set in memory. The reformulated compressive learning framework illustrates that compression is only possible for modest dimensions. 
\vspace*{-3.5mm}
\section{Conclusion}
\vspace*{-1.5mm}
Compressive learning for parametric models achieve successful compression in memory complexity as the sketch is commensurate with the model dimensions. In this paper, our case studies have shown that this is not always the case for semi-parametric models, as the identifiable statistic can scale well (ICA) or poorly (SC) with the underlying model dimensions. Importantly, our developed framework allows the user to identify exactly when memory compression is possible given an identifiable statistic, where the existing framework lacked. An interesting research direction which has arisen from this work is - ``Given an identifiable statistic associated with a semi-parametric model, is it of minimal dimensionality?".

\bibliographystyle{IEEEbib}
\bibliography{main}

\end{document}